\documentclass{article} 
\usepackage[final]{colm2026_conference}
\usepackage[T1]{fontenc}

\usepackage{microtype}
\usepackage{hyperref}
\usepackage{url}
\usepackage{booktabs}
\usepackage{soul}
\usepackage{graphicx}
\usepackage{booktabs}
\usepackage{multirow}
\usepackage{amsmath}
\usepackage{amsthm}

\usepackage[most]{tcolorbox}
\usepackage{listings}
\usepackage{xcolor}

\lstdefinestyle{promptstyle}{
  basicstyle=\ttfamily\footnotesize,
  breaklines=true,
  columns=fullflexible,
  keepspaces=true,
  showstringspaces=false
}

\newtcolorbox{promptbox}[2][]{
  enhanced,
  breakable,
  colback=gray!3,
  colframe=black!60,
  boxrule=0.5pt,
  arc=2pt,
  left=6pt,
  right=6pt,
  top=6pt,
  bottom=6pt,
  title={#2},
  fonttitle=\bfseries,
  #1
}

\usepackage{lineno}

\definecolor{darkblue}{rgb}{0, 0, 0.5}
\hypersetup{colorlinks=true, citecolor=darkblue, linkcolor=darkblue, urlcolor=darkblue}
\newtheorem{definition}{Definition}

\title{CT-$\Delta$Bench: A Benchmark for Longitudinal 3D Medical Imaging Difference Reporting with  Vision-Language Models}

\author{Kegeng Tang, Jingbo Wang, Shaogang Ren \& Zihao Wang\thanks{Corresponding author.} \\
University of Tennessee at Chattanooga\\
\texttt{\{hcw575,ydk297,sren9,zihao.wang\}@tennessee.edu} 
}

\begin{document}

\ifcolmsubmission
\linenumbers
\fi

\maketitle

\begin{abstract}
In medical imaging, the clinical value of Computed Tomography (CT) lies not only in depicting current disease status, but crucially in enabling longitudinal comparison of serial scans to determine disease evolution, a process that underpins response assessment, recurrence detection, and ongoing patient management. Yet, despite this central role of temporal comparison in clinical decision-making, existing medical foundation models remain largely confined to single-study understanding, leaving temporally grounded cross-examination insufficiently addressed. To address this gap, we study longitudinal imaging difference reporting, a task in which a model takes two temporally separated scans from the same patient and generates a clinically meaningful report describing interval changes between them. We introduce CT-$\Delta$Bench, a dedicated benchmark for this task with patient-level splitting to prevent information leakage. To better evaluate this task beyond surface-level text similarity, we further develop change-aware metrics specifically designed to capture clinically meaningful longitudinal changes, and conduct an independent physician validation to assess the reliability of the synthesized references and event extraction pipeline. We also compare direct paired-CT reasoning with an indirect two-stage pipeline that first generates single-timepoint reports and then performs textual differencing. Finally, we propose DeltaMed, a baseline model for direct paired-CT difference reporting, and train it on the benchmark training set. Together, these contributions lay the groundwork for temporally aware medical foundation models that better reflect real-world longitudinal clinical reasoning.\footnote{Code and data available at \url{https://github.com/LaplaceLab/CT-DeltaBench}}
\end{abstract}

\section{Introduction}

Medical imaging plays a central role in modern clinical care, with modalities such as X-ray, ultrasound, magnetic resonance imaging (MRI), and computed tomography (CT) providing complementary information for diagnosis, treatment planning, and disease monitoring \citep{hussain2022modern, zheng2026enhancing}. In many settings, their value lies not only in characterizing a single examination, but also in enabling longitudinal comparison across time \citep{zhou2025review}. Among these modalities, CT is particularly important for cancer surveillance, post-treatment assessment, and follow-up of chronic thoracic and progressive diseases, where clinicians often compare prior and current scans to determine whether abnormalities have newly appeared, progressed, regressed, resolved, or remained stable \citep{gai20253d, zheng2025corosam, wang2024ct}.

Despite this clinical importance, automated radiology report generation has so far focused mainly on single-study understanding. Existing CT-oriented report generation studies and 3D medical vision-language models still predominantly generate descriptions for one CT volume at a time, rather than directly reasoning over temporally paired scans \citep{bai2024m3d,chen20243dctgpt,hamamci2024ct2rep,sloan2024review}. However, follow-up interpretation depends not only on \emph{what is present}, but also on \emph{what has changed}, and such temporal judgments are often more clinically actionable than static descriptions alone. This problem is especially challenging for CT because paired-volume reasoning is computationally demanding, anatomical correspondence across time points is often imperfect, and many clinically important changes are subtle and localized \citep{yu2023evaluating}. In addition, when generating natural-language summaries of temporal differences, models are particularly vulnerable to omission and hallucination \citep{gai20253d,jain2021radgraph,yu2023evaluating}. As a result, longitudinal CT difference reporting is substantially more demanding than conventional single-study CT report generation.

To address this gap, we study \emph{longitudinal CT difference reporting}, in which a model receives two CT scans from the same patient and generates a clinically meaningful description of interval changes. We address this problem by establishing a dedicated benchmark, \textbf{CT-$\Delta$Bench}, with patient-level splitting to enable rigorous evaluation of paired-CT reasoning without information leakage. To better assess clinically meaningful temporal changes, we further develop change-aware evaluation metrics, motivated by the known limitations of conventional text-based metrics in radiology report generation \citep{delbrouck2022improving,ostmeier2024green}. Finally, we investigate both direct and indirect solution paradigms for this task, including direct paired-CT reasoning, indirect two-stage report differencing, and a dedicated baseline model, \textbf{DeltaMed}, trained on the benchmark training set.

The main contributions of this work are as follows: 1) We formulate \emph{longitudinal CT difference reporting} as a distinct benchmark task and introduce \textbf{CT-$\Delta$Bench}, a novel benchmark with patient-level splitting for systematic evaluation. 2) We develop change-aware evaluation metrics that better assess clinically meaningful longitudinal changes beyond surface-level text similarity. 3) We benchmark multiple large models under zero-shot and fine-tuning settings, including a controlled comparison between direct paired-CT reasoning and indirect two-stage pipelines. 4) We propose \textbf{DeltaMed}, a baseline 3D model for direct paired-CT difference reporting, and train it on the benchmark training set.

\section{Related Work}

\subsection{Single-study Medical Image Report Generation}

Automatic medical report generation has been studied primarily in radiology, especially for chest X-ray, where public datasets such as IU X-Ray, MIMIC-CXR, and CheXpert enabled large-scale development of image-to-text models \citep{demner2016preparing,johnson2019mimic,irvin2019chexpert}. Early approaches largely adapted generic image captioning architectures to produce reports from a single image or study, while later work emphasized longer-form generation, stronger visual--text alignment, improved factual consistency, and more clinically meaningful evaluation \citep{miura2021improving,nicolson2023improving}. More recently, this line of research has expanded to other medical imaging modalities, including CT, MRI, and volumetric settings, often supported by emerging medical vision-language models. Nevertheless, most existing methods still follow a \emph{single-study} formulation, in which the model generates a report for one exam at a time rather than explicitly reasoning over temporal relationships across multiple examinations from the same patient \citep{sloan2024review}.

\subsection{Longitudinal Medical Image Understanding}

A second line of work incorporates temporal context, prior studies, or longitudinal structure into medical vision--language learning, although most of it focuses on chest X-ray rather than volumetric imaging. For example, \cite{zhu2023utilizing} construct Longitudinal-MIMIC and use prior chest X-rays together with prior reports to improve current-report drafting; BioViL-T explicitly models temporal structure from image--report sequences and introduces MS-CXR-T for temporal biomedical vision--language evaluation \citep{bannur2023learning}; and MAIRA-2 incorporates realistic reporting context, including prior exams when available, into grounded report generation \citep{bannur2024maira}. More broadly, these studies show that longitudinal information can improve medical image understanding and report generation.

However, prior methods are still generally framed as current-exam reporting, where prior studies are used as auxiliary context, rather than as direct generation of a \emph{difference-aware} report whose primary purpose is to summarize interval change \citep{zhu2023utilizing,bannur2024maira,bannur2023learning}. In addition, most existing work remains centered on 2D chest X-rays rather than paired volumetric CT reasoning. In contrast, our benchmark focuses on direct paired-CT reasoning and explicit evaluation of clinically meaningful longitudinal change.

\subsection{Medical Benchmarks and Evaluation for Report Generation}

Public datasets and benchmarks have been central to medical report generation research. Classic resources such as IU X-Ray, MIMIC-CXR, and CheXpert supported much of the chest X-ray literature \citep{demner2016preparing,johnson2019mimic,irvin2019chexpert}, while more recent benchmarks such as CT-RATE, RadBench, and M3D-Bench broaden the scope toward 3D imaging, multimodal interaction, and general-purpose medical foundation models \citep{wu2025towards,bai2024m3d}. Despite this progress, most existing benchmarks still primarily evaluate single-study description, classification, retrieval, or question answering rather than explicit cross-timepoint difference reporting.

Evaluation has likewise become a major challenge. Traditional lexical metrics such as BLEU and ROUGE are easy to compute but often correlate poorly with clinical correctness. To address this, prior work has introduced more report-aware and fact-aware evaluation methods. CheXbert extracts structured labels from radiology reports for label-based assessment \citep{smit2020chexbert}, RadGraph defines a graph schema of entities and relations in radiology text \citep{jain2021radgraph}, and Yu et al.\ show that RadGraph F1 and RadCliQ correlate better with radiologist judgment than purely lexical metrics \citep{yu2023evaluating}. More recently, GREEN uses large language models to identify clinically significant report errors in a more interpretable way \citep{ostmeier2024green}. Collectively, these studies suggest that medical report evaluation should move beyond surface similarity toward factual and clinically meaningful correctness.

\section{Benchmark}
\subsection{Problem Definition}
\begin{figure}[t]
    \centering
    \includegraphics[width=0.9\textwidth]{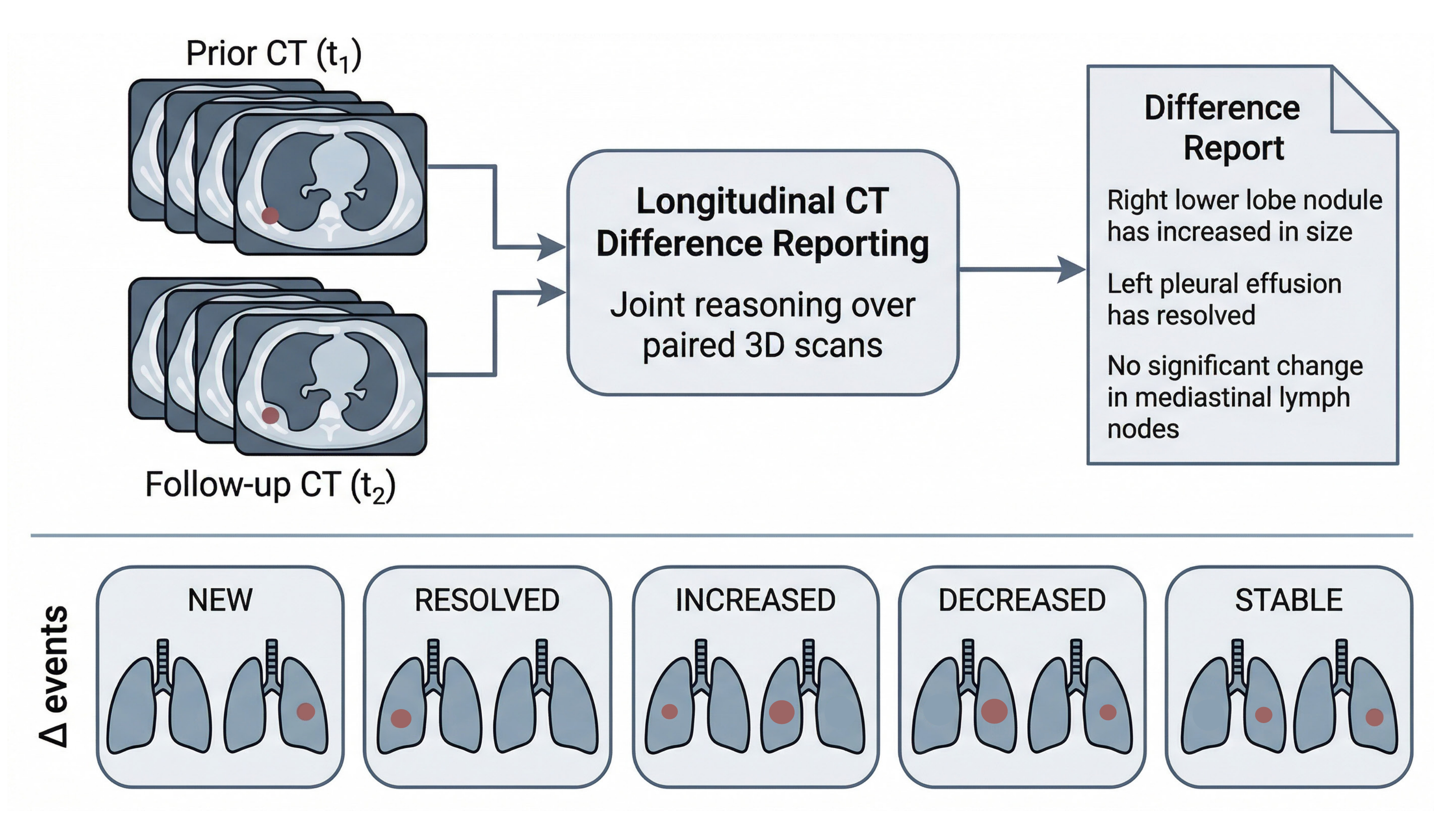}
    \caption{
    Task illustration of \emph{longitudinal CT difference reporting}. Given a prior CT scan $I_{t_1}$ and a follow-up CT scan $I_{t_2}$ from the same patient, the model generates a difference-aware report $R_{\Delta}$ describing clinically meaningful interval changes, such as new, resolved, increased, decreased, or stable findings.
    }
    \label{fig:task_definition}
\end{figure}

We study \emph{longitudinal CT difference reporting}, a benchmark task that evaluates whether a model can reason over two CT scans acquired from the same patient at different time points and generate a clinically meaningful report describing interval changes (Fig.~\ref{fig:task_definition}). Formally, each sample consists of a paired input $(I_{t_1}, I_{t_2})$, where $I_{t_1}$ and $I_{t_2}$ denote the earlier and follow-up CT scans, respectively. The target output is a difference-aware report $R_{\Delta}$ that summarizes clinically relevant temporal changes between the two studies, such as newly appeared findings, progression, regression, resolution, or stability.

Compared with conventional single-study CT report generation, this task requires explicit cross-timepoint reasoning. The benchmark is designed to evaluate not only whether a model can describe each scan individually, but also whether it can correctly identify and provide meaningful clinical changes description.

\subsection{Benchmark Construction}
We establish a longitudinal benchmark on top of CT-RATE \citep{hamamci2024ctrate}, a public dataset of 3D CT volumes and radiology reports. We identify patients with multiple CT studies and form longitudinal pairs by selecting two scans acquired at different time points for the same patient, denoted as an earlier scan $I_{t_1}$ and a follow-up scan $I_{t_2}$. Their corresponding radiology reports are denoted as $R_{t_1}$ and $R_{t_2}$.

CT-RATE provides report data from different time points, but does not include explicit clinically meaningful longitudinal descriptions. We construct a longitudinal target report for each paired sample using Gemini-2.5-Flash. Specifically, only the \textit{Findings} and \textit{Impression} sections from the prior and follow-up reports are used as model input. Given the source sections from $R_{t_1}$ and $R_{t_2}$, the model is instructed to generate a clinically grounded difference report that summarizes only the interval changes between the two studies in radiology-style natural language. The prompt explicitly encourages change-focused summarization while discouraging copying or exhaustively restating the full content of the original reports.

\begin{figure*}[t]
    \centering
    \includegraphics[width=\textwidth]{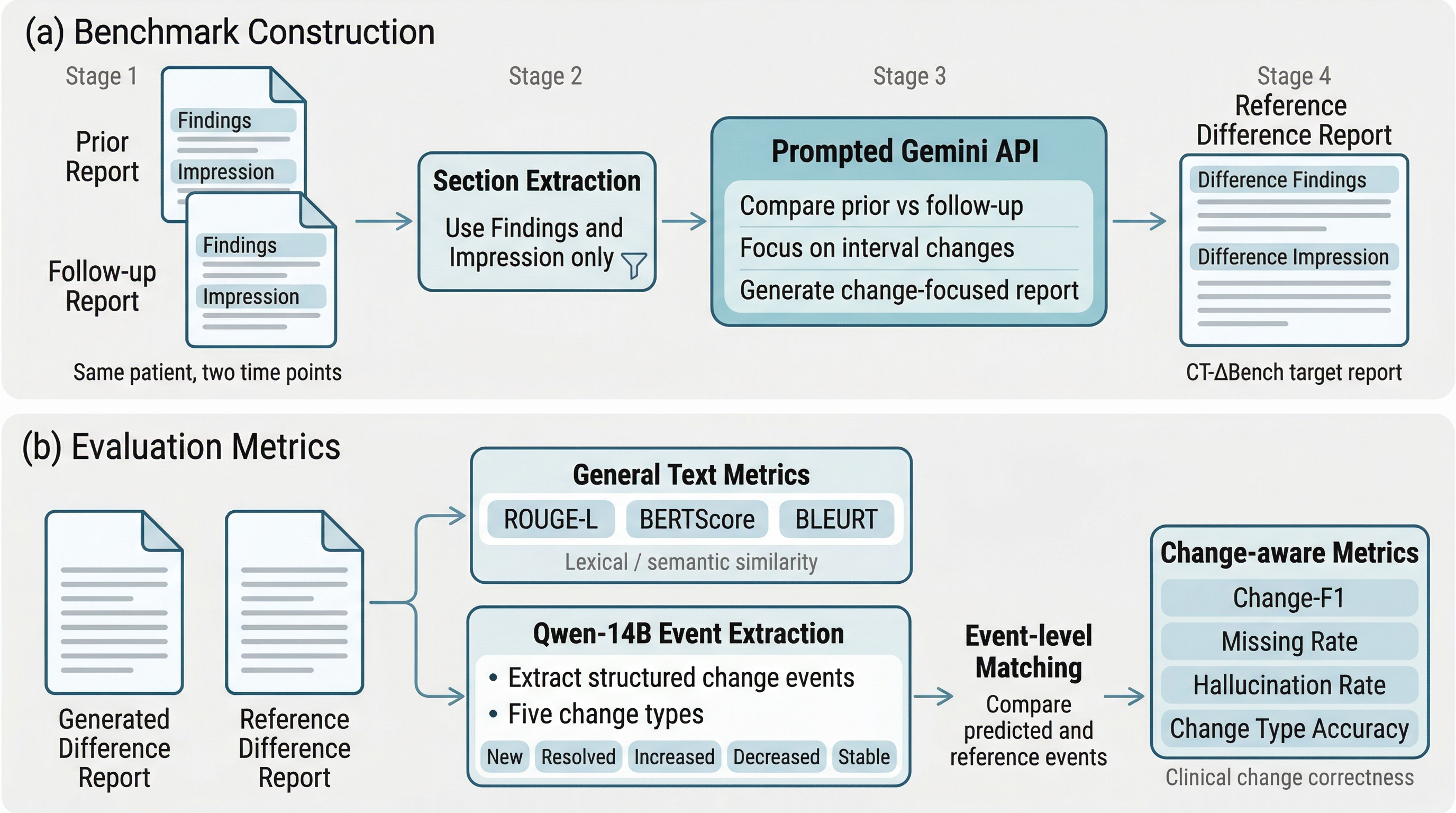}
\caption{Overview of benchmark construction and evaluation for CT-$\Delta$Bench. (a) Benchmark construction. For each longitudinal CT pair from the same patient, we retain only the \textit{Findings} and \textit{Impression} sections from the prior and follow-up radiology reports, and prompt Gemini-2.5-Flash to synthesize a reference difference report containing \textit{Difference Findings} and \textit{Difference Impression}. (b) Evaluation metrics. Model-generated difference reports are evaluated using both general text metrics (ROUGE-L, BERTScore, and BLEURT) and change-aware metrics. For change-aware evaluation, Qwen-14B is used to extract atomic change events with five change types (\textit{New}, \textit{Resolved}, \textit{Increased}, \textit{Decreased}, and \textit{Stable}), followed by event-level matching to compute Change-F1, Missing Rate, Hallucination Rate, and Change Type Accuracy.}
    \label{fig2}
\end{figure*}

As shown in Fig.~\ref{fig2}(a), the extracted report sections are passed to Gemini-2.5-Flash to generate a structured reference difference report consisting of \textit{Difference Findings} and \textit{Difference Impression}. The synthesized report is denoted as $R_{\Delta}$. It serves as the reference target for longitudinal CT difference reporting and focuses on temporal change description rather than single-timepoint report reconstruction. Each benchmark sample is therefore represented as a triple group: $(I_{t_1}, I_{t_2}, R_{\Delta})$, where the input is a paired CT study and the output is a difference-aware report. This construction enables systematic evaluation of models on longitudinal paired-CT reasoning and change-focused report generation.

The resulting benchmark is divided into training and validation sets for model training and evaluation. To prevent information leakage across subsets, we perform the split at the patient level. Specifically, each paired CT-report sample in the benchmark is drawn from a different patient. The training set contains 2,638 paired studies, and the validation set contains 169 paired studies. 

\subsection{Clinical Validation}

Because CT-$\Delta$Bench relies on LLM-synthesized reference difference reports
and LLM-based event extraction, we further conduct an independent clinical
validation to assess the reliability of these two components. We randomly sample
50 cases from the validation set and invite two physicians from different
hospitals to independently review the prior and follow-up CT reports, the
Gemini-synthesized difference reports, and the Qwen-extracted change events
using a structured rating form.

For the synthesized reference reports, the physicians assess overall
acceptability, correctness, and completeness on a five-point Likert scale, and
additionally determine whether each report is clinically acceptable and whether
it contains severe hallucinations or omissions. For the extracted change
events, they evaluate overall correctness and identify erroneous events or
missed important events.

As shown in Table~\ref{tab:clinical_validation}, the synthesized reference
reports achieve average scores of 4.82/5 for overall acceptability, 4.83/5 for
correctness, and 4.84/5 for completeness. Among the 100 physician evaluations,
99 are judged clinically acceptable, with no severe hallucination or severe
omission. The Qwen-based event extraction achieves an average correctness score
of 4.83/5, with only 3/100 evaluations identifying erroneous events and 3/100
identifying missed important events.

We further measure inter-physician consistency. The two physicians achieve
97.5\% positive-rating agreement (195/200), where both physicians assigning a score of 4 or 5 is counted as agreement, across the Likert-scale assessments
and 97.2\% agreement (243/250) across the binary clinical judgments. These
results provide targeted clinical evidence that the synthesized references and
event extraction pipeline are sufficiently reliable for benchmark-scale
evaluation.

\begin{table}[t]
\centering
\small
\caption{Clinical validation on 50 randomly sampled validation cases.
Two physicians from different hospitals independently evaluated each case.}
\label{tab:clinical_validation}
\begin{tabular}{lccc}
\toprule
Metric & Physician 1 & Physician 2 & Overall \\
\midrule
\multicolumn{4}{l}{\textit{Reference difference report}} \\
Overall acceptability & 4.86 & 4.78 & 4.82 \\
Correctness            & 4.88 & 4.78 & 4.83 \\
Completeness           & 4.86 & 4.82 & 4.84 \\
Clinically acceptable  & 50/50 & 49/50 & 99/100 \\
Severe hallucination   & 0/50  & 0/50  & 0/100 \\
Severe omission        & 0/50  & 0/50  & 0/100 \\
\midrule
\multicolumn{4}{l}{\textit{Change-event extraction}} \\
Overall correctness    & 4.84 & 4.82 & 4.83 \\
Erroneous event        & 3/50  & 0/50  & 3/100 \\
Missed important event & 0/50  & 3/50  & 3/100 \\
\bottomrule
\end{tabular}
\end{table}

\subsection{Evaluation Protocol}

As illustrated in Fig.~\ref{fig2}(b), we evaluate longitudinal CT difference reporting from two complementary perspectives: \emph{text-level quality} and \emph{event-level change correctness}. Since clinically valid difference reports may vary substantially in wording while describing the same temporal findings, text-based metrics alone are insufficient \citep{jain2021radgraph, yu2023evaluating}. We therefore report both conventional generation metrics and a structured event-based evaluation.

\paragraph{Text evaluation metrics.}
To measure overall textual similarity between a model-generated difference report $\widehat{R}_{\Delta}$ and the reference report $R_{\Delta}$, we report three standard text-generation metrics: ROUGE-L \citep{lin2004rouge}, BERTScore \citep{zhang2019bertscore}, and BLEURT \citep{sellam2020bleurt}. ROUGE-L measures longest common subsequence overlap and reflects surface-level lexical similarity. BERTScore evaluates semantic similarity based on contextual token embeddings. BLEURT further provides a learned quality score that is generally more robust to paraphrasing and wording variation. These metrics capture fluency and semantic alignment at the report level, but they do not explicitly assess whether the predicted report correctly identifies the clinically meaningful interval changes.

\paragraph{Event evaluation metrics.}
To directly evaluate whether a generated report correctly captures clinically meaningful interval changes, we propose an event-based evaluation protocol that compares extracted \emph{change events} rather than surface text alone. Following the change-aware evaluation pipeline in Fig.~\ref{fig2}(b), we use Qwen2.5-14B-Instruct to extract temporal events from both the generated report $\widehat{R}_{\Delta}$ and the reference report $R_{\Delta}$. Each report is converted into a set of atomic events, where each event is represented as a change label paired with a short free-text description of the corresponding finding. Concretely, each event follows a simple \texttt{(type, text)} format, where \texttt{text} denotes the finding mention and \texttt{type} specifies its temporal status. We use five clinically interpretable change categories: \texttt{NEW}, \texttt{RESOLVED}, \texttt{INCREASED}, \texttt{DECREASED}, and \texttt{STABLE}, corresponding respectively to lesion emergence, disappearance, progression, regression, and no substantial interval change. The extracted event sets are then used for event-level matching and for computing the four change-aware metrics described below.

Let $E_{\mathrm{pred}}$ and $E_{\mathrm{ref}}$ denote the event sets extracted from the predicted and reference reports, respectively. Under the fuzzy event-matching rule described in Appendix~\ref{app:event_matching}, we compute
\begin{equation}
\mathrm{TP}:=\lvert E_{\mathrm{pred}} \cap E_{\mathrm{ref}} \rvert,\qquad
\mathrm{FP}:=\lvert E_{\mathrm{pred}} \setminus E_{\mathrm{ref}} \rvert,\qquad
\mathrm{FN}:=\lvert E_{\mathrm{ref}} \setminus E_{\mathrm{pred}} \rvert,
\label{eq:event_counts}
\end{equation}
following the standard precision--recall formulation \citep{manning2008introduction}, where $\mathrm{TP}$, $\mathrm{FP}$, and $\mathrm{FN}$ denote the numbers of true-positive, false-positive, and false-negative events, respectively. Accordingly, we have
$\mathrm{Precision}:=\frac{\mathrm{TP}}{\mathrm{TP}+\mathrm{FP}}$ and
$\mathrm{Recall}:=\frac{\mathrm{TP}}{\mathrm{TP}+\mathrm{FN}}$.

\begin{definition}[Task-specific event-level metrics]
We define the event-level \emph{Hallucination Rate} as
\begin{equation}
\mathrm{Hallucination\ Rate}
:=
\frac{\mathrm{FP}}{\lvert E_{\mathrm{pred}} \rvert}
=
\frac{\mathrm{FP}}{\mathrm{TP}+\mathrm{FP}}.
\label{eq:hallucination_rate}
\end{equation}

Let $N_{\mathrm{type\mbox{-}correct}}$ denote the number of matched events whose predicted change type agrees with the reference change type. We further define the \emph{Change Type Accuracy} as
\begin{equation}
\mathrm{Change\ Type\ Accuracy}
:=
\frac{N_{\mathrm{type\mbox{-}correct}}}{\mathrm{TP}}.
\label{eq:change_type_acc}
\end{equation}
\end{definition}

In addition, we report two conventional event-detection metrics, namely the \emph{Change-F1} score and the \emph{Missing Rate}, given by
\begin{equation}
\mathrm{Change\mbox{-}F1}
:=
\frac{2\mathrm{TP}}{2\mathrm{TP}+\mathrm{FP}+\mathrm{FN}}\quad, \quad
\mathrm{Missing\ Rate}
:=
\frac{\mathrm{FN}}{\lvert E_{\mathrm{ref}} \rvert}
=
\frac{\mathrm{FN}}{\mathrm{TP}+\mathrm{FN}}.
\label{eq:missing_rate}
\end{equation}

At the event level, Change-F1 quantifies the overall agreement between predicted and reference change events. Missing Rate captures the proportion of reference events that are omitted in the prediction, whereas Hallucination Rate reflects the proportion of predicted events that are not supported by the reference. Change Type Accuracy measures whether the temporal semantic status assigned to a matched event is consistent with the reference annotation.

\subsection{Proposed Baseline Model}

\begin{figure}[t]
    \centering
    \includegraphics[width=\linewidth]{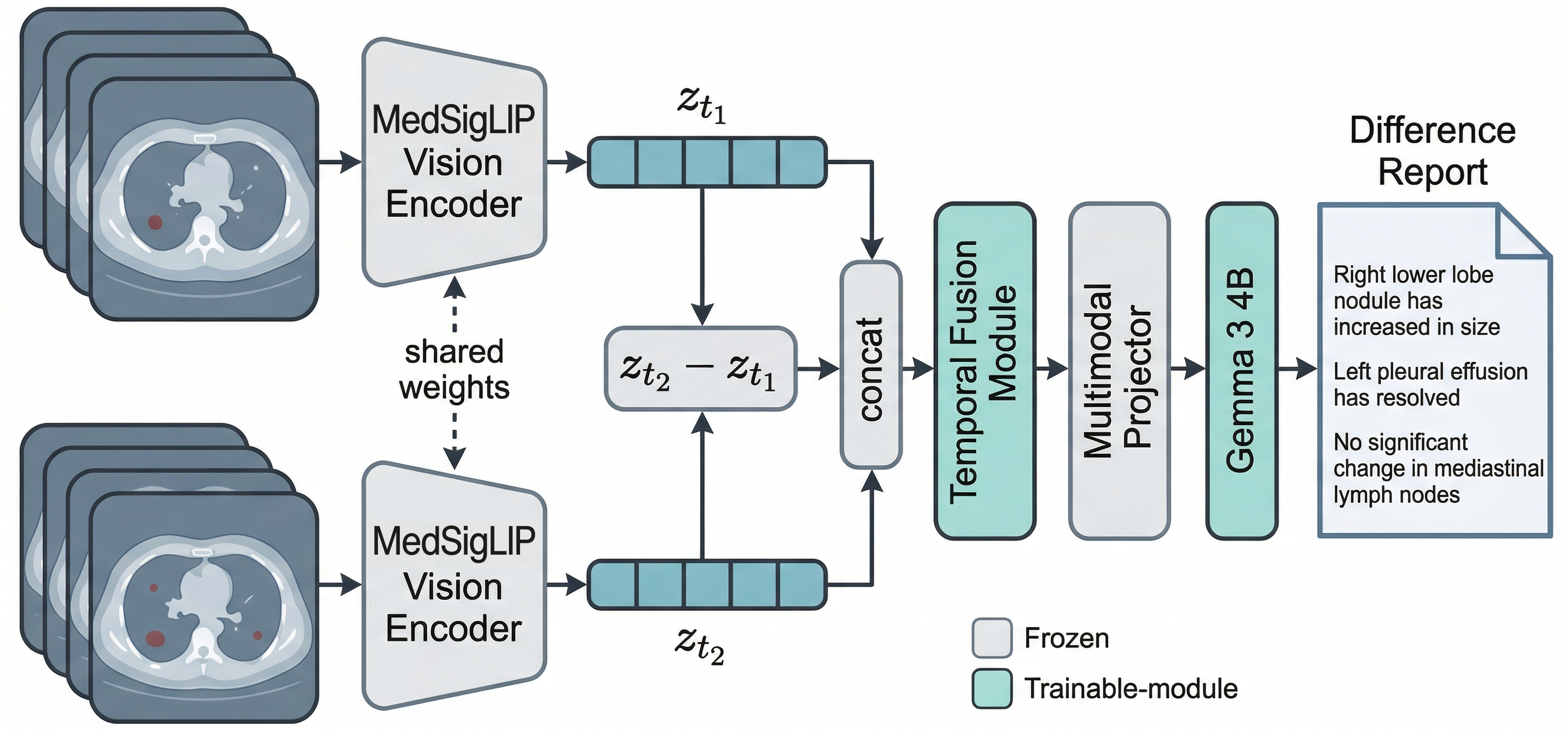}
\caption{Architecture of DeltaMed for longitudinal CT difference reporting. A prior CT and a follow-up CT are encoded by two MedSigLIP vision encoders with shared weights to obtain $z_{t_1}$ and $z_{t_2}$. A difference branch $z_{t_2}-z_{t_1}$ explicitly models temporal change. The resulting features are concatenated, fused, and passed through a multimodal projector and Gemma 3 4B to generate the final difference report. Frozen and trainable components are shown in gray and green, respectively.}
    \label{fig:deltamed}
\end{figure}

To address the gap in existing work on longitudinal CT difference reporting, we introduce DeltaMed, a dual-branch vision-language framework that explicitly models temporal change between a prior CT study and a follow-up CT study. Given a paired input $(I_{t_1}, I_{t_2})$, where $I_{t_1}$ denotes the prior CT and $I_{t_2}$ denotes the follow-up CT, DeltaMed first encodes the two time points separately using a shared visual encoder. Concretely, each scan is processed by the same MedSigLIP vision encoder \citep{sellergren2025medgemma}, producing two visual representations, $z_{t_1}$ and $z_{t_2}$. Sharing the encoder weights ensures that the two studies are mapped into a consistent feature space while avoiding unnecessary parameter growth.

To explicitly capture longitudinal change, DeltaMed constructs a difference-aware representation based on both scan-specific and temporal difference features. After obtaining $z_{t_1}$ and $z_{t_2}$, we compute an additional difference branch, $z_{t_2} - z_{t_1}$, which is intended to encode directional interval change from the prior study to the follow-up study. The three feature streams, $z_{t_1}$, $z_{t_2}$, and $z_{t_2} - z_{t_1}$, are then concatenated and passed through a lightweight temporal fusion module consisting of a linear projection followed by a normalization layer. This produces a fused longitudinal representation for downstream report generation.

The fused visual features are subsequently passed through the original multimodal projector and then fed into a Gemma 3 4B language model to generate the final difference report. In this way, DeltaMed performs direct joint reasoning over paired CT studies rather than relying on two independently generated single-study reports followed by textual differencing.


\noindent\textbf{Training Objective:} Let $H$ denote the fused longitudinal visual representation that encodes temporally evolving evidence across the prior and follow-up CT studies, and let $Y=(y_1,\ldots,y_T)$ denote the target difference-report sequence. We cast report generation as a conditional autoregressive decoding process and train the model by minimizing the sequence-level negative log-likelihood:
\begin{equation}
\mathcal{L}_{\mathrm{gen}}
=
-\sum_{t=1}^{T}
\log P(y_t \mid y_{<t}, H).
\label{eq:gen_loss}
\end{equation}
To preserve pretrained knowledge and reduce training cost, we adopt a parameter-efficient fine-tuning strategy: only the temporal fusion module and LoRA adapters inserted into the language model are updated, while the MedSigLIP vision encoder, the original multimodal projector, and the base Gemma 3 4B \citep{gemma32025} weights remain frozen. 

\section{Experiments}

\subsection{Experimental Setup and Baselines}

We conduct experiments on CT-$\Delta$Bench, a benchmark for longitudinal CT difference reporting introduced in this work. Each sample consists of a paired input $(I_{t_1}, I_{t_2})$ from the same patient, where $I_{t_1}$ denotes the prior CT and $I_{t_2}$ denotes the follow-up CT, together with a reference difference report $R_{\Delta}$. To prevent subject leakage across data partitions, we split the benchmark at the patient level into training and validation sets. In this work, zero-shot evaluation is primarily conducted on the validation set, while supervised fine-tuning uses the training split under different data regimes.

We consider three experimental settings. First, we benchmark five existing medical vision-language models in the zero-shot setting to evaluate their out-of-the-box ability on longitudinal CT difference reporting. The evaluated models include MedGemma-1.5-4B \citep{sellergren2025medgemma}, M3D-LaMed-Phi-3-4B \citep{bai2024m3d}, RadFM-13B \citep{wu2025towards}, Med3DVLM-Qwen2.5-7B \citep{xin2025med3dvlm}, and Merlin-RadLLaMA-7B \citep{blankemeier2026merlin}. Since most of these models are not specifically designed or optimized for jointly processing two CT studies as input, we directly feed each model with the paired CT studies in a zero-shot manner to assess its ability to perform longitudinal difference reporting without task-specific adaptation. Second, using the same set of models, we study a two-stage pipeline in which each model first generates an individual report for the prior CT and the follow-up CT separately. The resulting two single-study reports are then provided as textual input to the language model component of the same model, which is tasked with generating the final difference report. Third, we evaluate supervised fine-tuning under three training-data regimes, 1\%, 10\%, and 100\%, by applying LoRA to both DeltaMed and a direct paired-CT MedGemma baseline. All experiments are conducted on two 80GB NVIDIA A100 GPUs.

We report both text-level and event-level metrics. Specifically, we use ROUGE-L, BERTScore, and BLEURT to measure lexical and semantic similarity between generated reports and reference reports, and use Change-F1, Missing Rate, Hallucination Rate, and Change Type Accuracy to directly assess whether a model correctly captures clinically meaningful temporal changes. This combined evaluation protocol is necessary because multiple clinically valid difference reports may use different wording to describe the same interval changes, so text-level similarity alone cannot fully reflect clinical correctness. For all zero-shot experiments, we use a unified task instruction asking the model to generate a clinically meaningful difference report focused on interval changes only, without any in-context demonstrations. For supervised experiments, models are fine-tuned on the CT-$\Delta$Bench training split and evaluated on the same validation set using the identical metric suite.

\subsection{Results and Analysis}

\begin{table*}[t]
\begin{center}
\fontsize{6.5}{6.5}\selectfont
\setlength{\tabcolsep}{3pt}
\begin{tabular}{lccccccc}
\toprule
\multicolumn{1}{c}{\bf MODEL} &
\multicolumn{1}{c}{\bf ROUGE-L$\uparrow$} &
\multicolumn{1}{c}{\bf BERTScore$\uparrow$} &
\multicolumn{1}{c}{\bf BLEURT$\uparrow$} &
\multicolumn{1}{c}{\bf Change-F1$\uparrow$} &
\multicolumn{1}{c}{\bf Missing R$\downarrow$} &
\multicolumn{1}{c}{\bf Hallucination R$\downarrow$} &
\multicolumn{1}{c}{\bf Change Type Acc.$\uparrow$} \\
\midrule
MedGemma-1.5-4B            &0.0754 &0.7931 &0.3611 &0.0175 &0.9849 &0.9791 &0.5909 \\
M3D-LaMed-Phi-3-4B         &0.0890 &0.7986 &0.3706 &0.0051 &0.9966 &0.9899 &0.4000 \\
RadFM-13B                  &0.0537 &0.7597 &0.3157 &0.0000 &1.0000 &1.0000 &0.0000 \\
Med3DVLM-Qwen2.5-7B        &0.0980 &0.7964 &0.3822 &0.0138 &0.9918 &0.9562 &0.0000 \\
Merlin-RadLLaMA-7B         &0.0705 &0.8059 &0.3493 &0.0034 &0.9979 &0.9897 &1.0000 \\

\bottomrule
\end{tabular}
\end{center}
\caption{Zero-shot performance of different models on CT-$\Delta$Bench. Standard text-generation metrics (ROUGE-L, BERTScore, and BLEURT) are reported together with change-aware metrics, including Change-F1, Missing Rate (R), Hallucination Rate (R), and Change Type Accuracy.}
\label{tab:zeroshot_benchmark}
\end{table*}

\paragraph{Zero-shot benchmarking of existing models.}
Table~\ref{tab:zeroshot_benchmark} shows that all evaluated models perform poorly on the proposed benchmark in the zero-shot setting, especially on the change-aware metrics. Across all models, Change-F1 remains extremely low, ranging only from $0$ to $0.0175$. In particular, RadFM-13B completely fails to recover matched change events, with Change-F1 of $0$, Missing Rate of $1$, and Hallucination Rate of $1$. Even the best event-level result, achieved by MedGemma-1.5-4B, yields only a Change-F1 of $0.0175$, together with a Missing Rate of 0.9849 and a Hallucination Rate of $0.979$. These numbers indicate that current models rarely identify clinically meaningful interval changes correctly, while also frequently generating unsupported change statements.

The results also reveal a clear disconnect between text-level similarity and temporal change correctness. Med3DVLM-Qwen2.5-7B achieves the best ROUGE-L ($0.098$) and BLEURT ($0.3822$), and Merlin-RadLLaMA-7B achieves the best BERTScore ($0.8059$), yet their change-aware metrics remain poor. In particular, Med3DVLM-Qwen2.5-7B reaches only $0.0138$ Change-F1, while Merlin-RadLLaMA-7B falls to $0.0034$ despite its strongest BERTScore. This contrast suggests that conventional text-generation metrics may overestimate performance for longitudinal difference reporting, where correct identification of temporal changes is more important than surface-level textual similarity. Change Type Accuracy should also be interpreted jointly with event matching quality. For instance, Merlin-RadLLaMA-7B obtains Change Type Accuracy of $1$, but this occurs together with Change-F1 of only $0.0034$, indicating that the apparently strong type accuracy is supported by very few matched events. Overall, these findings highlight a substantial gap between existing zero-shot medical vision-language models and the clinical demands of longitudinal CT difference reporting.

\begin{table*}[t]
\begin{center}
\fontsize{6.5}{6.5}\selectfont
\setlength{\tabcolsep}{3pt}
\begin{tabular}{lccccccc}
\toprule
\multicolumn{1}{c}{\bf MODEL} &
\multicolumn{1}{c}{\bf ROUGE-L$\uparrow$} &
\multicolumn{1}{c}{\bf BERTScore$\uparrow$} &
\multicolumn{1}{c}{\bf BLEURT$\uparrow$} &
\multicolumn{1}{c}{\bf Change-F1$\uparrow$} &
\multicolumn{1}{c}{\bf Missing R$\downarrow$} &
\multicolumn{1}{c}{\bf Hallucination R$\downarrow$} &
\multicolumn{1}{c}{\bf Change Type Acc.$\uparrow$} \\
\midrule
MedGemma-1.5-4B            & 0.0886 & 0.7822 & 0.3303 & 0.0078 & 0.9952 & 0.9790 & 0.4286 \\
M3D-LaMed-Phi-3-4B         & 0.0448 & 0.2957 & 0.3635 & 0.0212 & 0.9877 & 0.9250 & 0.3889 \\
RadFM-13B                  & 0.0753 & 0.8159 & 0.3545 & 0.0542 & 0.9603 & 0.9147 & 0.2414 \\
Med3DVLM-Qwen2.5-7B        & 0.1450 & 0.8213 & 0.3579 & 0.0614 & 0.9500 & 0.9204 & 0.2055 \\
Merlin-RadLLaMA-7B         & 0.0803 & 0.8006 & 0.3841 & 0.0000 & 1.0000 & 1.0000 & 0.0000 \\
\bottomrule
\end{tabular}
\end{center}
\caption{Performance of the two-stage difference reporting pipeline on CT-$\Delta$Bench. In this setting, each model first generates separate reports for the prior CT and the follow-up CT, and then uses the two generated single-study reports as textual input to produce the final difference report.}
\label{tab:twostage_benchmark}
\end{table*}

\paragraph{Two-stage difference reporting based on separately generated CT reports.}
We further evaluate a two-stage difference reporting pipeline in which each model first generates separate reports for the prior CT and the follow-up CT, and then uses the two reports as textual input to produce the final difference report. As shown in Table~\ref{tab:twostage_benchmark}, the two-stage pipeline yields mixed results compared with direct zero-shot paired-CT prompting in Table~\ref{tab:zeroshot_benchmark}. The largest gains are observed for RadFM-13B and Med3DVLM-Qwen2.5-7B, both of which show improved Change-F1 together with lower Missing Rate and Hallucination Rate. M3D-LaMed-Phi-3-4B also improves on event-level metrics, although its text-level scores become less stable. In contrast, MedGemma-1.5-4B becomes slightly worse on change-aware metrics, and Merlin-RadLLaMA-7B degrades most severely, with Change-F1 dropping to $0.0000$. Overall, these results suggest that indirect textual differencing can help some models recover temporal change information, but the benefit is inconsistent and remains inferior to reliable grounded paired-image reasoning. This inconsistent behavior is likely due to error propagation from the intermediate single-study reports. In the two-stage setting, the final difference report can only compare findings preserved in the first-stage reports. If a single-study report omits a finding, the corresponding interval change becomes unrecoverable; conversely, hallucinated findings may be amplified into spurious changes during textual differencing. This explains why models producing more informative single-study reports, such as RadFM-13B and Med3DVLM-Qwen2.5-7B, benefit more from the two-stage pipeline, whereas models with noisier or less complete intermediate reports may degrade.

\begin{table*}[t]
\begin{center}
\fontsize{6.5}{6.5}\selectfont
\setlength{\tabcolsep}{3pt}
\begin{tabular}{lcccccccc}
\toprule
\multicolumn{1}{c}{\bf MODEL} &
\multicolumn{1}{c}{\bf Regime} &
\multicolumn{1}{c}{\bf ROUGE-L$\uparrow$} &
\multicolumn{1}{c}{\bf BERTScore$\uparrow$} &
\multicolumn{1}{c}{\bf BLEURT$\uparrow$} &
\multicolumn{1}{c}{\bf Change-F1$\uparrow$} &
\multicolumn{1}{c}{\bf Missing R$\downarrow$} &
\multicolumn{1}{c}{\bf Hallucination R$\downarrow$} &
\multicolumn{1}{c}{\bf Change Type Acc.$\uparrow$} \\

\midrule
\multirow{3}{*}{MedGemma}
& 1\%   & 0.0716 & 0.7791 & 0.3430 & 0.0010 & 0.9993 & 0.9984 & \underline{1.0000} \\
& 10\%  & 0.1724 & \underline{0.8196} & 0.3458 & 0.0649 & 0.9110 & 0.9489 & \underline{0.7538} \\
& 100\% & 0.2825 & \underline{0.8665} & \underline{0.4063} & 0.1577 & 0.8856 & \underline{0.7462} & 0.4671 \\
\midrule
\multirow{3}{*}{DeltaMed}
& 1\%   & \underline{0.1348} & \underline{0.7978} & \underline{0.3549} & \underline{0.0909} & \underline{0.9288} & \underline{0.8744} & 0.5385 \\
& 10\%  & \underline{0.1831} & 0.8028 & \underline{0.3803} & \underline{0.1313} & \underline{0.9093} & \underline{0.8565} & 0.5424 \\
& 100\% & \underline{0.2899} & 0.8558 & 0.4006 & \underline{0.1980} & \underline{0.8301} & 0.8057 & \underline{0.4902} \\

\bottomrule
\end{tabular}
\end{center}
\caption{Performance of DeltaMed and the direct paired-CT MedGemma baseline under different supervised fine-tuning regimes on CT-$\Delta$Bench. Both methods are evaluated at 1\%, 10\%, and 100\% training-data regimes.}
\label{tab:finetune_regimes}
\end{table*}

\paragraph{DeltaMed and direct paired-CT MedGemma under different fine-tuning regimes.}
We further compare DeltaMed with a direct paired-CT MedGemma-1.5-4B baseline under three supervised fine-tuning regimes using 1\%, 10\%, and 100\% of the training set, with LoRA applied in all cases. As shown in Table~\ref{tab:finetune_regimes}, both models improve as more training data become available, but DeltaMed consistently achieves better event-level change detection across all regimes. In the 1\% regime, DeltaMed improves Change-F1 from $0.001$ to $0.091$ while reducing Missing Rate from $0.999$ to $0.929$ and Hallucination Rate from $0.998$ to $0.874$. In the 10\% regime, DeltaMed still outperforms MedGemma on Change-F1 ($0.1313$ vs.\ $0.0649$), Missing Rate ($0.9093$ vs.\ $0.9110$), and Hallucination Rate ($0.8565$ vs.\ $0.9489$). Under full-data fine-tuning, DeltaMed remains better on Change-F1 ($0.1980$ vs.\ $0.1577$) and Missing Rate ($0.8301$ vs.\ $0.8856$), although MedGemma attains a lower Hallucination Rate. Text-level metrics are more mixed: DeltaMed achieves higher ROUGE-L in all three regimes, whereas MedGemma becomes slightly stronger in BERTScore and BLEURT at higher-data regimes; however, these gains do not translate into better change-aware performance. Change Type Accuracy should also be interpreted jointly with event matching quality, since MedGemma reaches $1$ in the 1\% regime despite a Change-F1 of only $0.001$. Overall, these results suggest that DeltaMed provides a stronger inductive bias for longitudinal change understanding than a generic direct paired-CT adaptation of MedGemma, especially in low-data settings.

\section{Conclusion}

In this work, we introduce CT-$\Delta$Bench, a dedicated benchmark for longitudinal CT difference reporting, a clinically important yet underexplored task that requires models to reason over paired CT studies and generate reports describing interval changes. To support this setting, we construct a patient-level benchmark based on longitudinal CT pairs, develop change-aware evaluation metrics that go beyond surface-form text similarity, and systematically benchmark existing medical vision-language models under both direct paired-CT and indirect two-stage settings. Our results show that current models remain far from solving this task, especially when evaluated on clinically meaningful event-level change correctness rather than only text-level similarity. We further propose DeltaMed, a simple baseline that explicitly models temporal difference through paired-CT reasoning and a difference branch. Experimental results show that DeltaMed provides stronger event-level change detection than a direct paired-CT MedGemma baseline across multiple fine-tuning regimes, particularly when supervision is limited. Overall, our study establishes a clearer task formulation, a reproducible evaluation framework, and a strong baseline for future research on temporally aware medical foundation models for longitudinal clinical imaging.

\section*{Acknowledgments}

This research is supported by the Ruth S. Holmberg Grant, and is supported in part by the faculty development grant from the University of Tennessee at Chattanooga. We thank Xiao Xiao from Chongqing Medical University and Zhidu Wang from The Third Bethune Hospital of Jilin University for their valuable assistance with the clinical validation of CT-$\Delta$Bench. We also thank the anonymous reviewers for their constructive feedback.

\section*{Ethics Statement}

CT-$\Delta$Bench uses LLM-synthesized, report-derived reference reports and is intended solely for controlled research evaluation rather than clinical deployment. Although our physician validation provides a targeted assessment of reference and event-extraction quality, any clinical use would require larger-scale prospective expert validation of both the benchmark references and model outputs.

\bibliography{colm2026_conference}
\bibliographystyle{colm2026_conference}

\clearpage
\appendix
\section{Appendix}
\subsection{Prompt Template}
\label{app:prompts}

We use Gemini 2.5 Flash with the \emph{Longitudinal Difference Report Synthesis Prompt} to synthesize longitudinal differential report data from paired original CT report JSONs. The source data consist of two CT reports from the same patient at different time points (\texttt{Report A} as prior and \texttt{Report B} as current), using only the \texttt{Findings\_EN} and \texttt{Impressions\_EN} fields. The target data are structured longitudinal comparison reports in JSON format, containing \texttt{patient\_id}, \texttt{VolumeName\_A}, \texttt{VolumeName\_B}, \texttt{Findings\_EN}, and \texttt{Impressions\_EN}.

\begin{promptbox}{Longitudinal Difference Report Synthesis Prompt}
\label{app:synthesis}
\begin{lstlisting}[style=promptstyle]
You are a radiology longitudinal report synthesizer. Use ONLY the provided two original CT report JSON objects and focus ONLY on their Findings_EN and Impressions_EN fields. Do NOT use ClinicalInformation_EN, Technique_EN, any label vectors, or external medical knowledge. Do NOT invent details that are not stated.

INPUT:
- Report A JSON (prior / earlier exam): {REPORT_A_JSON}
- Report B JSON (current / later exam): {REPORT_B_JSON}

PATIENT ID RULE:
Derive "patient_id" automatically from VolumeName by taking the substring before the last two underscore-separated tokens.
Example: "train_10006_a_1.nii.gz" -> "train_10006"

TASK:
1) Parse VolumeName_A and VolumeName_B from the two JSON inputs.
2) Treat Report A as PRIOR and Report B as CURRENT. Write a differential (interval-change) summary describing B relative to A.
3) Base every statement strictly on Findings_EN and Impressions_EN text from the two reports.
4) Do not use or summarize ClinicalInformation_EN or Technique_EN.
5) If a finding is only mentioned in one report, describe it as "mentioned/not mentioned" rather than assuming true absence/presence clinically.
6) When content overlaps, explicitly state whether it is unchanged/stable between reports.
7) Keep the writing concise, clinically phrased, and change-focused.

OUTPUT:
Return ONE valid JSON object ONLY (no markdown, no extra text), with EXACTLY these keys:
- "patient_id"
- "VolumeName_A"
- "VolumeName_B"
- "Findings_EN"
- "Impressions_EN"

FIELD GUIDELINES:
- Findings_EN: Write one coherent paragraph based only on Findings_EN and Impressions_EN content: start with stable findings (airways, mediastinum/lymph nodes, pleura/pericardium, cardiovascular, lungs), then list interval differences/newly mentioned items. Mention important additions in the current report as "newly documented/mentioned".
- Impressions_EN: 2-4 sentences summarizing overall interval change and the most clinically relevant newly documented items vs stable findings.

Now generate the output JSON.
\end{lstlisting}
\end{promptbox}

We use Qwen2.5-14B-Instruct with the \emph{Longitudinal Change Event Extraction Prompt} to convert the original longitudinal CT difference reports (\texttt{Findings\_EN} + \texttt{Impressions\_EN}) into structured target data consisting of atomic change events labeled as \texttt{NEW}, \texttt{RESOLVED}, \texttt{INCREASED}, \texttt{DECREASED}, or \texttt{STABLE}.

\begin{promptbox}{Longitudinal Change Event Extraction Prompt}
\begin{lstlisting}[style=promptstyle]
You are an information extractor for longitudinal CT DIFFERENCE reports.

INPUT
- patient_id: {PATIENT_ID}
- DELTA_REPORT_TEXT (Findings + Impression ONLY):
{DELTA_REPORT_TEXT}

OUTPUT (JSON ONLY)
Return exactly:
{
  "patient_id": "<patient_id>",
  "events": [
    {"type": "NEW|RESOLVED|INCREASED|DECREASED|STABLE", "text": "<short event>"},
    ...
  ]
}

RULES
1) Use ONLY DELTA_REPORT_TEXT. No external knowledge. Do NOT invent details.
2) Extract ATOMIC change events (one finding/change per event). Split combined sentences.
3) Allowed types ONLY: NEW, RESOLVED, INCREASED, DECREASED, STABLE.
4) Remove ALL numbers and measurements from "text" (mm/cm/% and any digits).
5) If the change type/direction is not explicit or not confident, OMIT the event (no "uncertain/indeterminate").
6) Do not output items that are only "not mentioned" or "limited evaluation" without a clear change.
7) Keep "text" concise, clinically phrased, no section headers, no duplicates.

Now output the JSON only.

\end{lstlisting}
\end{promptbox}

\subsection{Data Cases}

The following example illustrates a synthesized longitudinal difference report in CT-$\Delta$Bench. It is represented as a JSON object containing the patient identifier, paired volume names, and difference-aware \texttt{Findings\_EN} and \texttt{Impressions\_EN} fields.
\begin{promptbox}{Longitudinal Difference Report Example}
\begin{lstlisting}[style=promptstyle]
{
  "patient_id": "train_2163",
  "VolumeName_A": "train_2163_a_1.nii.gz",
  "VolumeName_B": "train_2163_b_1.nii.gz",
  "Findings_EN": "The trachea and main bronchi, mediastinal lymph nodes, heart and mediastinal vascular structures, esophagus, pleural spaces, bilateral adrenal glands, and abdominal and bone structures were mentioned as natural or within normal limits in the prior report but are not explicitly described in the current report. In the current examination, the peripherally arranged and round-looking ground glass-like densities previously observed in almost all areas of both lungs are noted to be reduced in volume, fainter, and observed in a more amorphous morphology. The crazy paving appearances and consolidations noted in the prior report are not explicitly mentioned in the current report. Interstitial scars are again evident.",
  "Impressions_EN": "The prior report suggested viral pneumonia, possibly COVID, with a differential diagnosis. The current examination's findings are evaluated in favor of regression of the previously noted ground glass densities. No explicit impression is given in the current report."
}

\end{lstlisting}
\end{promptbox}

The following example illustrates the event extraction output used for change-aware evaluation. Each longitudinal difference report is converted into a JSON object containing the patient identifier and a list of atomic change events, where each event is represented by a change \texttt{type} and a short free-text \texttt{text} description.
\begin{promptbox}{Longitudinal Change Event Extraction Example}
\begin{lstlisting}[style=promptstyle]

{
  "patient_id": "valid_403",
  "events": [
    {
      "type": "INCREASED",
      "text": "Size of mediastinal lymph nodes"
    },
    {
      "type": "NEW",
      "text": "Ground-glass opacities in upper zones and lower lobes"
    },
    {
      "type": "NEW",
      "text": "Mild emphysema"
    },
    {
      "type": "NEW",
      "text": "Hepatosteatosis"
    },
    {
      "type": "RESOLVED",
      "text": "Pneumonic consolidations"
    },
    {
      "type": "STABLE",
      "text": "Bronchial wall thickening"
    },
    {
      "type": "INCREASED",
      "text": "Number of subpleural blebs"
    },
    {
      "type": "NEW",
      "text": "Sequelae changes at apical levels"
    },
    {
      "type": "DECREASED",
      "text": "Density in liver"
    },
    {
      "type": "INCREASED",
      "text": "Degenerative changes in bone structures"
    }
  ]
}
\end{lstlisting}
\end{promptbox}

\subsection{Fuzzy Event Matching Details}
\label{app:event_matching}

To support event-level evaluation, we align the reference and predicted event sets by \texttt{patient\_id} and then perform fuzzy event matching between the two sets. Since predicted and reference events may use different but semantically similar phrasings, exact string matching is often too brittle for longitudinal difference reporting. We therefore adopt a normalized matching procedure that combines text canonicalization, hard clinical constraints, and soft similarity scoring.

\paragraph{Text canonicalization.}
For each event text, we first apply canonicalization to reduce superficial lexical variation. Specifically, we lowercase the text, apply phrase-level synonym normalization (e.g., \emph{ground-glass opacities} $\rightarrow$ \emph{ground glass opacity}), and perform token-level normalization such as plural-to-singular conversion and common lexical variant normalization. This step is intended to make semantically equivalent event descriptions more directly comparable.

\paragraph{Clinical constraint filtering.}
From the canonicalized event text, we further extract simple clinical cues, including laterality labels (e.g., \emph{left}, \emph{right}, \emph{bilateral}) and coarse anatomy tags (e.g., \emph{lung}, \emph{pleura}, \emph{mediastinum}, and \emph{lymph node}). A candidate reference--prediction pair is rejected if it exhibits a hard conflict, namely disjoint laterality labels or disjoint anatomy tags. This step reduces clearly implausible matches before soft similarity is computed.

\paragraph{Soft similarity scoring and one-to-one matching.}
For all remaining candidate pairs, we compute a soft textual similarity score using token-level F1 on the canonicalized event texts. Pairs with similarity below a threshold of $\tau = 0.5$ are discarded. We then enforce a one-to-one matching between reference and predicted events by maximizing the number of valid matches, using total similarity as a secondary criterion. For large event sets, a greedy approximation is used for efficiency. Matched pairs are treated as aligned events for subsequent evaluation, while unmatched reference and predicted events are treated as misses and spurious predictions, respectively.

\paragraph{Change type comparison.}
After event matching is established, change labels are compared only within the matched event pairs. In this way, change type correctness is evaluated conditioned on successful event alignment, rather than on the full unmatched event sets.

\end{document}